\documentclass[a4paper]{document-classes/llncs}

\usepackage[right=2cm,top=2cm,bottom=2cm,left=2cm]{geometry}

\usepackage{amsmath}
\usepackage{amssymb}
\usepackage{setspace}
\usepackage{enumitem}
\usepackage{booktabs}
\usepackage[title]{appendix}
\usepackage{url}
\usepackage{graphicx}
\usepackage{caption, subcaption}
\usepackage{xspace}
\usepackage{tabularx}
\usepackage{multicol}
\usepackage{changepage}

\usepackage{algorithm2e}
\SetKwComment{Comment}{/* }{ */}
\RestyleAlgo{ruled}
\usepackage{algpseudocode}

\usepackage{bm}

\usepackage{latexsym}
\usepackage{multirow}
\usepackage{lipsum}
\usepackage{xspace}
\usepackage[colorlinks=true,allcolors=blue]{hyperref}
\usepackage{mathrsfs}

\title{One Strike, You’re Out: Detecting Markush Structures in Low Signal-to-Noise Ratio Images}

\author{Thomas Jurriaans\inst{1}\inst{2}\quad
    Kinga Szarkowska\inst{2}\quad
    Eric Nalisnick\inst{1}\\
    Markus Schw\"{o}rer\inst{3}\quad
    Camilo Thorne\inst{3}\quad
    Saber Akhondi\inst{2}
}
\institute{
    University of Amsterdam, Amsterdam\\
    \href{mailto:thomasjurriaans@gmail.com}{\tt thomasjurriaans@gmail.com} (corresponding) \quad \email{e.t.nalisnick@uva.nl}
    \and
    Elsevier BV, Amsterdam\\
    \email{k.szarkowska@elsevier.com} \quad \email{s.akhondi@elsevier.com}
    \and
    Elsevier Information Systems GmbH, Frankfurt\\
    \email{c.thorne.1@elsevier.com} \quad \email{m.schwoerer@elsevier.com}
}

\begin{document}

\maketitle

\begin{abstract}
Modern research increasingly relies on automated methods to assist researchers. An example of this is Optical Chemical Structure Recognition (OCSR), which aids chemists in retrieving information about chemicals from large amounts of documents. Markush structures are chemical structures that can not be parsed correctly by OCSR and cause errors. The focus of this research was to propose and test a novel method for classifying Markush structures. Within this method, a comparison was made between fixed-feature extraction and end-to-end learning (CNN). The end-to-end method performed significantly better than the fixed-feature method, achieving 0.928 $\pm$ 0.035 Macro F1 compared to the fixed-feature method's 0.701 $\pm$ 0.052. Because of the nature of the experiment, these figures are a lower bound and can be improved further. These results suggest that Markush structures can be filtered out effectively and accurately using the proposed method. When implemented into OCSR pipelines, this method can improve their performance and use to other researchers.

\vspace{0.1cm}
\href{https://github.com/Thomasjurriaans/markush-recognition-msc-thesis}{\noindent{\bf GitHub Repository.}} [Code, demos and public datasets.]
\end{abstract}


\section{Introduction}
\label{sec:introduction}

As the amount of publications increases over time~\cite{larsen2010rate}, scientists require more sophisticated and automated methods to keep overview of the published research in their field of study. While search engines such as \href{https://scholar.google.com}{Google Scholar} and \href{https://www.semanticscholar.org}{Semantic Scholar} might suffice for most fields, in chemistry searching for relevant works can be more difficult. Chemists might search for very specific molecules that can not easily be named, or for properties that certain molecules possess. It is for this reason that Optical Chemical Structure Recognition (OCSR) is used to convert Chemical Structural Formulae (CSF) from 2D image representations to machine-readable formats \cite{rajan2020review}. This allows for the creation of chemical databases, in which a user can easily find all publications related to a specific molecule. An example of such a database is Elsevier's Reaxys \cite{elsevierreaxys}. 

One problem with using OCSR is the presence of Markush structures \cite{csepregi2011representation}. They are CSF's that are incomplete and function as a "template". Performing OCSR on these gives false outputs\cite{khokhlov2022image2smiles}, so the OCSR performance can be improved by filtering them out.

\begin{figure}[ht!]
    \centering
    \begin{subfigure}[t]{0.45\textwidth}
        \centering
        \includegraphics[width=0.5\textwidth]{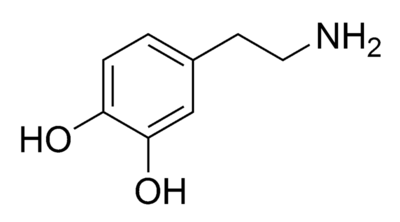}
   \caption{A complete CSF.}
    \end{subfigure}
    ~~
    \begin{subfigure}[t]{0.45\textwidth}
        \centering
        \includegraphics[width=0.5\textwidth]{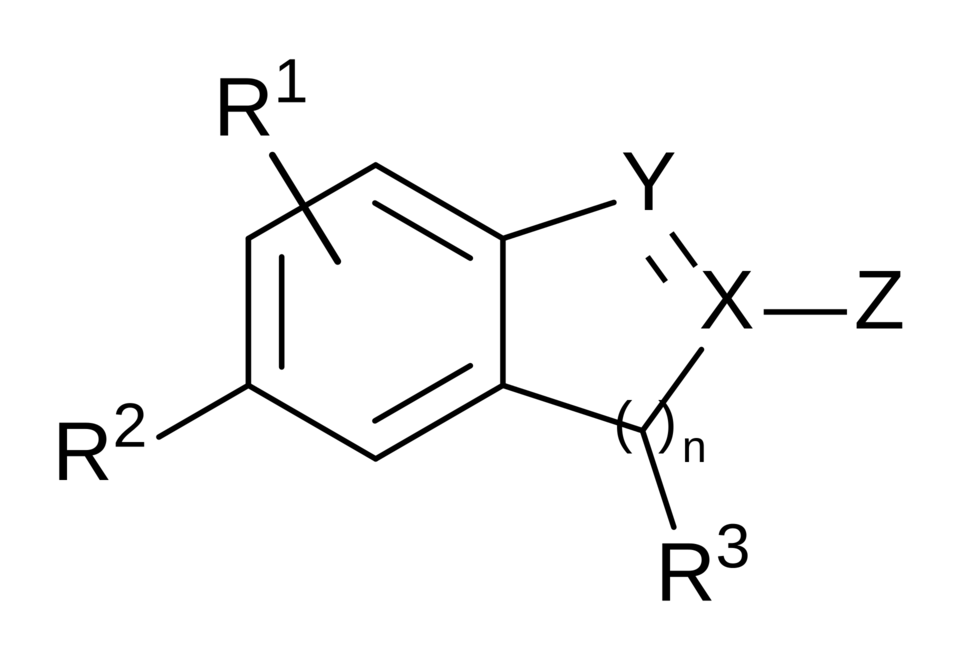}
        \caption{An incomplete (Markush) CSF. The R's, X, Y, Z and $()_n$ are all Markush indicators.}
    \end{subfigure}
    \caption{}
    \label{fig:markush_example} 
\end{figure}

Markush structures can be recognized by the use of a variety of symbols and structures in a CSF, we will call these ``Markush indicators". There are no standard guidelines for drawing Markush structures, even experienced chemists sometimes struggle to recognize them \cite{khokhlov2022image2smiles}. Apart from this, recognition of images containing Markush structures comes with a variety of other challenges. Challenging enough that some other researchers chose to manually exclude Markush structures from their datasets, even though these structures are still present in the real-world dataset \cite{xu2022molminer}. The reason for this is that there is no standardized format in which scientific works are published. More specifically, CSF's are almost always presented in 2D images, which may be of any size, contain one or multiple CSF's and can also contain other figures or text. Furthermore, the pixel scale also differs per image, a particular Markush indicator might be 5 pixels wide in one image, and 25 pixels wide in another. Moreover, the presence of only one small Markush indicator in the entire image means that a Markush structure is present. This changes the classification of the entire image, hence the title of this paper. This means that the image has low Region of Interest (ROI)-to-Image ratio \cite{kong2022efficient}. This results in a low Signal-to-Noise Ratio (SNR), which makes training harder. This problem is exacerbated by the cost of labeling data: this can only be done by, or with supervision of, chemistry experts, meaning that large amounts of labeled data are not available or very expensive.

The goal of this research is to determine the efficacy of a proposed novel method to solve these problems in Markush classification. Furthermore, it aims to determine whether end-to-end Deep Learning (DL) is required to solve this problem, or whether a fixed-feature extraction method can achieve similar results. This work therefore aims to answer the following question:
\begin{quote}
    \textit{To what extent can end-to-end deep convolutional neural networks improve classification of low signal-to-noise ratio images of Markush structures compared to fixed-feature computer vision methods?}    
\end{quote}
with the following sub-question:
\begin{quote}
    \textit{To what extent are the features used to determine the classification interpretable?  And if they are interpretable, do they contain reliable explanatory power?}
\end{quote}
\section{Related Work}
\label{sec:related_work}

Research in the field of image classification is mostly focused on images with a large amount of information and large ROI-to-image ratios. Examples of such images can be found in ImageNet \cite{deng2009imagenet}, likely the most popular and widely used image dataset. Furthermore, even datasets used for few-shot image classification contain many more training examples than the dataset available for this problem \cite{liu2022few}. This makes it hard to compare these results to the problem this work is aimed to solve. Kong and Henao in \cite{kong2022efficient} propose a method that is able to efficiently classify large images with very small ROI's. Their method uses an end-to-end CNN model that, crucially, does not require pixel-level annotations. It does this by making use of an attention-based method with two stages, in which the first stage determines areas of interest and the second stage only processes those areas. However, this method is not applicable to our problem, as it requires that all input images are of equal dimensions. Furthermore, almost all modern works use deep convolutional networks for classification, and do not make a comparison to more traditional fixed-feature computer vision (CV) methods. 

Research focused on OCSR deals with Markush structures, but not with the same goal as this work aims to do. For example, Xu et al. in \cite{xu2022molminer} perform OCSR using deep learning, but choose to exclude Markush structures from their dataset. While Khokhlov et al. in \cite{khokhlov2022image2smiles} do process Markush structures, however they convert them directly to a machine-readable format (FG-SMILES, see \cite{weininger1988smiles}) instead of classifying them separately. Furthermore, their method uses images containing only a single structure, which was cropped manually. They report that many of the failures of their model were due to Markush structures, which substantiates that OCSR performance can be improved by classifying Markush structures and filtering them out.

In order to answer the main research question, a fixed feature extraction method will be compared to a Convolutional Neural Network. The ORB (Oriented FAST and Rotated BRIEF) feature detector \cite{rublee2011orb} will be used. With over 10000 citations it is one of the most widely used CV feature detectors and therefore a good representation of a "traditional" CV technique to compare to. 

ORB detects features in an image, and encodes them as a string of bits. Features can be compared by comparing each bit individually: the more bits that are different, the less similar the features are. The amount of bits that differ between two features is called the Hamming distance. In addition to the ORB feature extraction, Lowe's ratio test \cite{Lowe_2004} is used to filter out good matches. The ratio test checks that the two Hamming distances between the first and second-best keypoint matches are sufficiently different. This helps to filter out ambiguous matches (where the distance ratio between the two nearest neighbors is close to one) from well discriminated matches.

The conclusion that can be drawn from these related works is that classifying and filtering Markush structures in OCSR pipelines can be valuable for improving performance. However, no methods have been published that solve the unique issues that come with this specific image classification task.
\section{Methodology}
\label{sec:methodology}

To answer the research questions, analytical experimentation was conducted by creating 2 image processing pipelines. A thorough description of the data used as well as the pipelines created is given in the following section.

\subsection{Datasets}

\subsubsection{Primary Dataset}

The data in the primary dataset consists of images with CSF's in them. Some of the images contain multiple CSF's and some also include other structures such as text or tables. These images are representative of how images would be extracted from the scientific publications that Elsevier has access to. These images were manually labeled by a domain expert, in 3 different classes:
\begin{enumerate}[label=\Alph*:]
    \item Complete substances (108 images)
    \item Images containing at least 1 Markush structure (122 images)
    \item Images that contain only Markush structures. (42 images)
\end{enumerate}

For this project, if an image contained at least one Markush structure, it is to be filtered out. This means that classes B and C could be combined. In this project, an image containing no Markush structure is considered "false", and an image containing at least one Markush structure is considered "true". More information on the primary dataset can be found in \autoref{table:Dataset_stats}.

\begin{figure}[tb]
    \centering
    \includegraphics[width=0.30\textwidth]{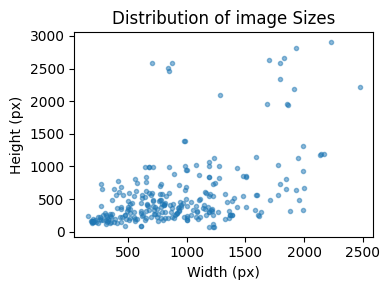}
    \caption{Scatter plot show the image size distribution of the primary dataset.}
    \label{fig:ImageSizeDistribution} 
\end{figure}

\subsubsection{Auxiliary Dataset}

During the later stages of the project, a secondary dataset was made available by Elsevier. This dataset contains 5117 complete structure images and 317 images containing at least one Markush structure. This dataset was used to further evaluate the proposed methods.

\subsection{Dataset Problems} \label{sec:dataset_problems}

There are problems with the dataset which make classification more difficult. These issues can be summarized as follows:

\begin{itemize}
    \item \textbf{Low Signal-to-Noise ratio}\\
    The images contain very little signal because they have a small ROI. Most of the image is irrelevant for classification: only the symbols that indicate a Markush structure are relevant.
    \item \textbf{Large Image Size Variance}\\
    Some images are small (e.g. 160x240px) while others are large (1200x1000px), this means that standard approaches such as cropping or scaling to make all images of equal size are not feasible. This would cut out too much information or distort the image too much. The distribution of image sizes can be seen in \autoref{fig:ImageSizeDistribution}
    \item \textbf{Small Dataset}\\
    The dataset is not very large (272 images total), and some little-used symbols that indicate a Markush structure have only a few examples.
    \item \textbf{Pixel Scale Variance}\\
    The symbols and structures that need to be recognized differ in scale, one letter that indicates a Markush structure might be 5px wide in one image while being 25px wide in another.
    \item \textbf{Weak Annotation}\\
    The dataset is weakly annotated, meaning the images only have labels for the entire image. This is specifically an issue in combination with the low SNR and small dataset. With only weak annotation, we can not direct the classifier to focus on the relevant parts of the image.
\end{itemize}

\begin{figure}[tb]
    \centering
    \includegraphics[width=0.6\textwidth]{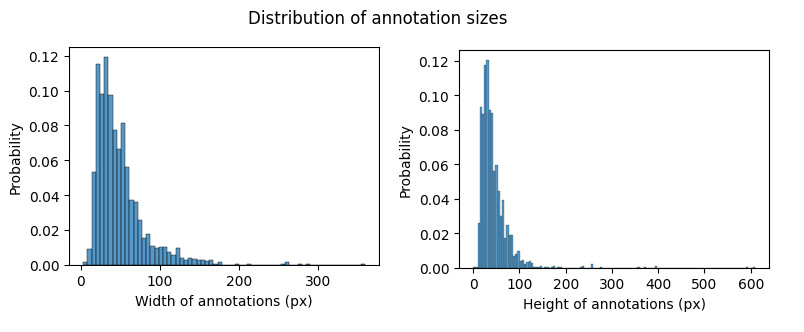}
   \caption{Histograms of annotation pixel sizes.}
    \label{fig:Annotations_hist} 
\end{figure}

\begin{figure}[tb]
    \centering
    \includegraphics[width=0.6\textwidth]{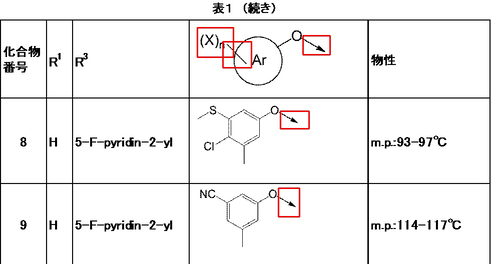}
    \caption{An example of a manually annotated image containing Markush structures in a Chinese patent (annotations in red).}
    \label{fig:annotations_example} 
\end{figure}

\subsection{Data Annotation}

As a means to resolve some of the issues mentioned in \autoref{sec:dataset_problems}, the relevant regions of the training images were manually annotated. This was done using supervision from the domain expert. This pixel-level annotation annotates which approximate area of the image is responsible for the classification as a Markush structure. This only needs to be done for images in categories 2 and 3. Since class 3 is a subset of class 2, we combine them into one class. This annotation was labor-intensive, but it was the only way to solve the ``SNR", ``small dataset" and "weak annotation" problems. Without more detailed annotation, there was simply too little information for the model to learn from. \autoref{fig:Annotations_hist} shows the size distribution of these annotations, and \autoref{fig:annotations_example} shows an example. More statistics of the dataset can be found in \autoref{table:Dataset_stats}. Only the primary dataset was annotated by hand, as the auxiliary dataset was only used for evaluation, no annotating was necessary.

\begin{table}[tb]
\centering
\begin{tabular}{l@{\qquad}l}
\# of images containing Markush           & 164 \\ 
\# of images not containing Markush       & 108  \\ 
Average \# of annotations per Markush image & 9.52   \\ 
Annotation width 99th quantile           & 160 px \\ 
Annotation height 99th quantile          & 174 px \\ 
Label ratio (Non-Markush:Markush)        & 40:60  \\ 
&
\end{tabular}
\caption{Primary dataset statistics.}
\label{table:Dataset_stats} 
\end{table}

\begin{figure}[tb]    
    \centering
    \begin{minipage}{0.45\textwidth}
    \centering
    	\includegraphics[width=0.7\textwidth]{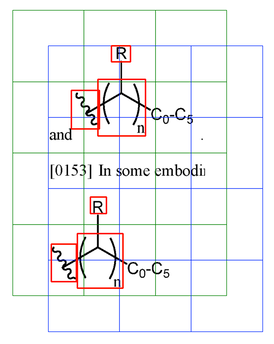}
    	\caption{An image containing Markush structures (red) is split into patches (green and blue) for classification by the CNN.}
    	\label{fig:patch_grid} 
    \end{minipage}
    ~~
    \begin{minipage}{0.45\textwidth}
    \centering
    	\includegraphics[width=0.82\textwidth]{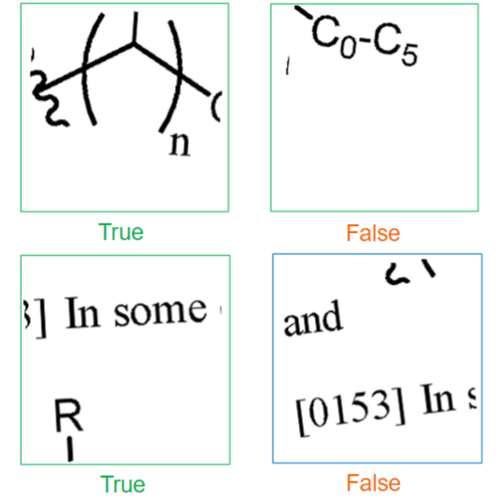}
    	\caption{4 out of the 24 patches generated from \autoref{fig:patch_grid} with their corresponding labels.}
   	 \label{fig:patch_gen} 
    \end{minipage}    
\end{figure}

\subsection{Patch generation}

In order to focus on the relevant parts of the images and address the "large image size variance issue", the images are cropped to create "patches". The CNNs are then trained on these patches. These patches are generated from 2 grids, as illustrated in \autoref{fig:patch_grid} and \autoref{fig:patch_gen}. The grids are offset by half a patches' width and height, this ensures that if an annotation falls on an edge in one grid, it will be centered in the other. The image is padded with white pixels because of this offset. Patches are cropped from the images and are labeled as either ``Markush" or ``Non-Markush" depending on whether they contain a Markush indicator or not. If an annotation is partially contained in a patch, it will be labeled ``Markush" if more than 50\% of the annotation's pixels fall inside the generated patch. Instead of classifying the entire image at once, the CNN was trained on these individual patches. During inference, the query image is broken up and the parches are individually classified by the CNN. The patches are 224x224 or 299x299 pixels, depending on the model architecture used. As seen in \autoref{fig:Annotations_hist}, this means that most annotations will fit completely inside the patch, meaning no information is lost. For the annotation width and height, the 99th percentile lies at 160 and 174 pixels respectively. For values above that, at least 99\% of annotations are expected to fit inside the patch with room to spare for context around the annotation.

Special "template" patches were also generated. These are patches that are centered on an annotation, and contain only the contents of the annotation, thus limiting noise. These are used for the ORB classification method where they function as a template.

\begin{figure}[tb]
    \centering
    \includegraphics[width=0.3\textwidth]{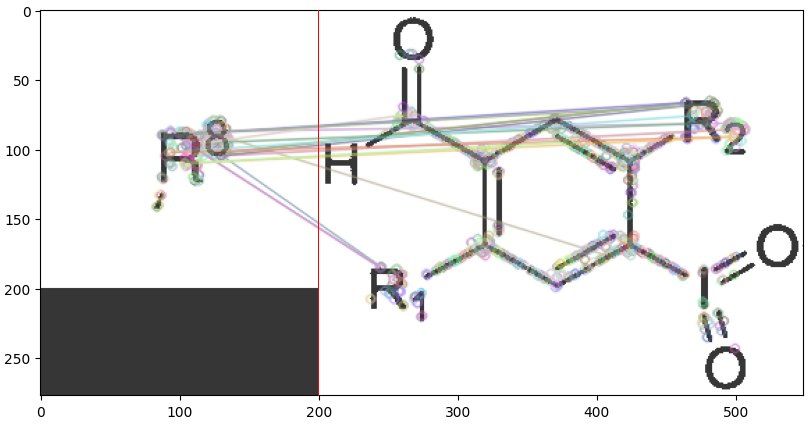}
   \caption{Keypoint matching between a Markush template patch (left) and a query image (right).}
    \label{fig:orb_comparison} 
\end{figure}

\subsection{Fixed-Feature Method (ORB)}

To answer the research question, a fixed-feature method needed to be compared to a deep learning approach. For this method, it was decided to use ORB \cite{rublee2011orb}. In order to perform classification with ORB, additional steps needed to be taken. ORB is a feature detector, and does not output a classification. Instead, it outputs keypoints for an image, which can then be compared with another image. An example is shown in \autoref{fig:orb_comparison}. 
This matching is done using a large amount of patches, each containing a different Markush indicator from the training set. After the comparison, we have the following information:

\begin{itemize}
\item How many keypoints each image has.
\item How many matches between these images there are.
\item What the Hamming distance is for each of these matches.
\item The pixel locations of the matching keypoints.
\end{itemize}

This numerical data was then used by a machine learning algorithm to create classifications. This algorithm can be as simple as logistic regression and does not require any deep learning. It was decided to use XGBoost \cite{xgboost} in this research. The reason for this decision is that XGBoost is a very widely used non-deep learning model that has been shown to perform well on a wide variety of problems, making it a good general choice.
The numerical data is generated as shown in \autoref{alg:1}.

\begin{algorithm}[]
\SetAlgoLined
\KwIn{$M$ Markush template patches, $N$ query images}
\KwOut{data frame $D$ with $N$ rows, $6M$ columns}

\For{each query image}{
\For{each template}{
calculate ORB matches between the query image and the current template\\
add to $D$:
\begin{itemize}
    \item the $M$ matches
    \item the Hamming distance of the 5 best matches
\end{itemize}
\vspace{-0.25cm}
}
}
\caption{Generating a DataFrame with ORB matches and distances}
\label{alg:1}
\end{algorithm}

This algorithm generates a DataFrame which is ready for any standard ML classifier. In this experiment, the data was split into a 60/20/20 train/validate/test split. A grid search was performed over the following hyperparameters:
\begin{itemize}
    \item The number of ORB features
    \item The number of template patches to compare to
    \item The number of XGB estimators
    \item The maximum depth of the XGB estimators
\end{itemize}

The hyperparameter grid that was searched is available in \autoref{tab:orb_hyperparameters}. It was kept small to keep computation times low, and because earlier tests showed only small performance differences between hyperparameter configurations. The hyperparameters that performed best on the validation set were evaluated on the test set. In this evaluation, the XGBoost model was retrained on both the train and validation sets, and evaluated on the test set, leading to the results as seen in \autoref{sec:results}.

\subsection{Deep Learning Method}

\subsubsection{Architectures}

For the Deep Learning method, two Convolutional Neural Network (CNN) architectures were compared. Specifically, the ResNet18 \cite{DBLP:journals/corr/HeZRS15} and InceptionV3 \cite{DBLP:journals/corr/SzegedyVISW15} architectures. They were implemented using PyTorch \cite{NEURIPS2019_Pytorch}. The ResNet series was chosen as it is a popular architecture, well-supported and widely used. This makes it a good representation of a deep learning CNN model.  Since the data is not highly complex, it was hypothesized that it was not necessary to use very deep architectures. It is for this reason that the ResNet18 architecture was chosen, as it is the shallowest of the ResNet models. 

For comparing with the ResNet18 model, the InceptionV3 architecture was chosen. The main reason for this was that Elsevier had an InceptionV3 Model that was pretrained on a large dataset containing only chemical images, which made for a valuable comparison. Furthermore, the InceptionV3 comparison allowed for the answering of one of the sub-research questions: whether a deeper model with more parameters would perform better.

\begin{table}[tb]
\centering
\begin{tabular}{c|c|c|c}
\multicolumn{1}{l|}{Model Architecture} & \multicolumn{1}{l|}{Input Size} & \multicolumn{1}{l|}{No. Layers} & \multicolumn{1}{l}{No. Parameters} \\
\hline
ResNet18 & 224x224 & 18 & 11.5 Million \\ \hline
InceptionV3 & 299x299 & 41 & 23.8 Million \\ 
\end{tabular}
\vspace{0.2cm}
   \caption{Architecture summary}
    \label{table:Architecture Summary} 
\end{table}

\subsubsection{Pretraining and fine-tuning}

\begin{figure}[tb]
    \centering
    \includegraphics[width=0.3\textwidth]{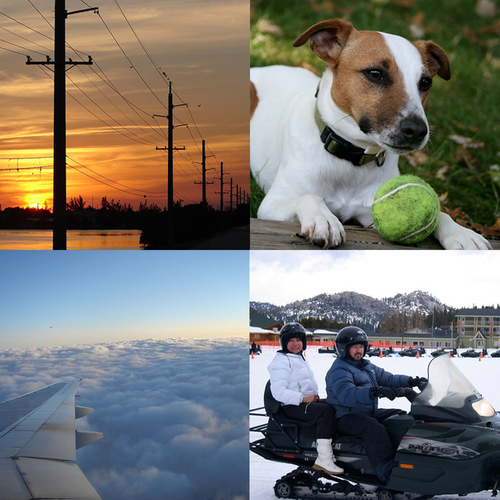}
   \caption{4 Sample images from the ImageNet pre-training dataset.}
    \label{fig:ImageNetSample} 
\end{figure}

\begin{figure}[tb]
    \centering
    \includegraphics[width=0.44\textwidth]{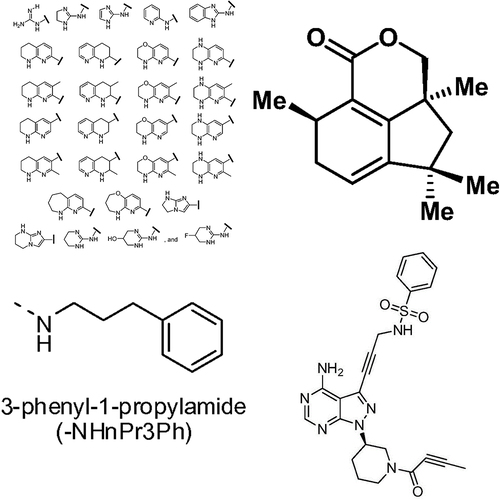}
   \caption{4 Sample images from the USPTO pre-training dataset.}
    \label{fig:USPTOSample} 
\end{figure}

Because of the small dataset, only pretrained CNNs were tested. Two pretraining datasets were compared, ImageNet and a USPTO dataset. ImageNet \cite{deng2009imagenet} is an image dataset that contains millions of images belonging to 1000 different classes. The images are all photos of real-world objects and organisms. The models used that were pretrained on ImageNet had the goal to classify to which of the 1000 classes an image belongs. The USPTO \cite{USPTO} dataset was compiled by Elsevier. It contains 10k images of CSF's and 10k images from scientific literature that do not contain CSF's. The goal of models pretrained on this dataset had the task of classifying images according to the two classes. While the pretraining tasks were equal (classification) the size of the data and the type of data is very different for the ImageNet and USPTO datasets. The ImageNet dataset has the advantage of a very large size, while the USPTO dataset has the advantage of being trained on the same domain. Example image of both datasets can be found in \autoref{fig:ImageNetSample} and \autoref{fig:USPTOSample}. 

The pre-trained models were fine-tuned on the primary dataset. This was done on a server with a Tesla V100-SXM2-16GB GPU and Intel Xeon E5-2686 CPU @ 2.30GHz and 64 GB of RAM. Two different variations of fine-tuning were tried. In the first variation, all layers were frozen except for the fully connected (FC) layers. This means that the low-level filters of the CNN are fully determined by the pretraining. In the second variation, no layers were frozen during training and thus the full model (FM) was trained. This gives the model the ability to change low-level features during training, at an increased risk of overfitting and a higher computational cost.

\subsubsection{Preprocessing and Augmentation}

As most CNNs require a 3-channel input (RGB), the single grayscale channel was duplicated 3 times. Although the input in each channel is exactly equal, the model's trained features differ per channel, allowing each channel to contribute uniquely to the classification. In order to improve performance and increase robustness, data augmentation was used. This artificially increases the equivalent size of the dataset. It was chosen to use augmentations that fit the domain, and could be found in (scanned) scientific documents:
\begin{itemize}
    \item random perspective shift,
    \item random posterization, and
    \item random sharpness or blur.
\end{itemize}
Each augmentation occurred with a pre-determined probability, which is a hyperparameter of the training.

\subsubsection{Hyperparameters, Training and Evaluation}

In this experiment, different models, pre-training options and fine-tuning methods were evaluated. Such a comparison can only be made fairly under otherwise equal circumstances. This raises the question on what hyperparameters to use. After all, the optimal hyperparameters will be different for each configuration, so it would not be fair to use the same hyperparameters for all. In this research it was therefore decided to use the Optuna \cite{DBLP:journals/corr/abs-1907-10902} library for hyperparameter optimization. 

For each configuration, Optuna finds the optimal hyperparameters using trials. A single trial means that the model is trained on the training set and evaluated on the validation set. The hyperparameters are different in each trial. In this way the trials are used to sample from the hyperparameter-performance distribution, and thus find where the optimum lies. It was chosen to sample 25 times for each experiment in order to balance performance and computational cost. At the end of the hyperparameter search, the model is trained on the merged training and validation sets using the optimal hyperparameters, and evaluated on the test set. Optuna was set to optimize for Macro F1 score, and this process was repeated 5 time for each model so a margin of error could be calculated. The hyperparameter ranges explored can be found in \autoref{tab:hyperparam_explored}.

The amount of epochs the model was trained for was not directly optimized by Optuna. Instead, early stopping was used. If the validation macro F1 score did not reach a new best score for $n$ epochs, training would be terminated. The optimal epoch count returned was always the one that returned the highest validation macro F1 score. The Optuna and PyTorch hyperparameters that were not optimized and thus consistent for each configuration can be found in \autoref{tab:optuna_pytorch_params}.

\subsubsection{Result Metrics}

With the classification of all the patches in the test set done, performance metrics can be calculated. In this research it was chosen to mainly use Macro F1 scores. The Macro F1 score can be calculated as follows: 
\begin{equation}
\text{{Macro F1}} = \frac{1}{N} \sum_{i=1}^{N} 2 \cdot    \frac{\text{{precision}}_i \cdot \text{{recall}}_i}{\text{{precision}}_i + \text{{recall}}_i}
\end{equation}
where $N$ is the amount of classes (2 in our case), and $\text{{recall}}_i$ and $\text{{precision}}_i$ are the recall and precision for class $i$. The values are averaged for each class equally. This makes the Macro F1 score reliable even when the class distribution is strongly skewed.

Each configuration was trained and evaluated 5 times. This allowed for calculating the margin of error of the result using the following formula: $\text{{Margin of Error}} = t_{\alpha/2, n-1} \cdot \frac{\sigma}{\sqrt{n}}$ with values
$\text{{Margin of Error}} = t_{0.025, 4} \cdot \frac{\sigma}{\sqrt{5}}$. Because of the small sample size of 5, the student T-distribution was used for the critical value. The commonly used value of $0.05$ for $\alpha$ was chosen, which gives a 95\% margin of error.

Since the model gives outputs for each patch that an image consists of, a label must still be computed for the entire image. This can be done by simply using this formula: $\text{X} = \max_{i=1}^{n} \left\{ x_i \right\}$ Where $X$ is the label of an image, $n$ is the amount of patches generated from $X$ and $x_i$ is the label of the $i$th patch in image $X$. This method simply labels the entire image as ``Markush" if any one of the patches is labeled as such. It would be possible to optimize this procedure further to improve the performance at the image-level, however this was outside the scope of this research.
\section{Results}
\label{sec:results}

\begin{table}[tb]
\begin{adjustwidth}{-1in}{-1in}
\centering
\small
\begin{tabular}{|lcccccccc|}
\hline
Configuration  & \multicolumn{1}{l}{Architecture} & \multicolumn{1}{l}{Input Size} & \multicolumn{1}{l}{Pretraining} & Layers     & \multicolumn{1}{l}{Patch - Val F1} & \multicolumn{1}{l}{Patch - Test F1} & \multicolumn{1}{l}{Image - Test F1} & \multicolumn{1}{l|}{Image - Aux Test F1} \\ \hline
ORB\_BASELINE  & ORB                              & Full Image                     & N.A                             & N.A        & N.A                                & N.A                                 & 0.701 $\pm$ 0.052                                & 0.533                                    \\
R18\_IN\_FC    & ResNet18                         & 224x224                        & ImageNet                        & FC only    & 0.803                              & 0.744 $\pm$ 0.022                       & 0.751 $\pm$ 0.094                       & 0.528                                    \\
R18\_IN\_FM    & ResNet18                         & 224x224                        & ImageNet                        & Full Model & 0.852                              & 0.889 $\pm$ 0.022                       & 0.839 $\pm$ 0.098                       & 0.816                                    \\
IV3\_USPTO\_FC & InceptionV3                      & 299x299                        & USPTO                           & FC only    & 0.596                              & 0.568 $\pm$ 0.071                       & 0.517 $\pm$ 0.100                       & 0.538                                    \\
IV3\_USPTO\_FM & InceptionV3                      & 299x299                        & USPTO                           & Full Model & 0.859                              & 0.759 $\pm$ 0.044                       & 0.759 $\pm$ 0.187                       & 0.863                                    \\
IV3\_IN\_FC    & InceptionV3                      & 299x299                        & ImageNet                        & FC only    & 0.725                              & 0.710 $\pm$ 0.030                       & 0.697 $\pm$ 0.140                       & 0.700                                    \\
IV3\_IN\_FM    & InceptionV3                      & 299x299                        & ImageNet                        & Full Model & \textbf{0.862}                     & \textbf{0.917 $\pm$ 0.014}              & \textbf{0.928 $\pm$ 0.035}              & \textbf{0.914}                           \\ \hline
\end{tabular}
   \vspace{0.2cm}
   \caption{Main experiment results.}
    \label{table:main_results} 
\end{adjustwidth}
\end{table}

\begin{table}[tb]
\centering
\begin{tabular}{|p{0.20\textwidth}|p{0.70\textwidth}|}
\hline
\centering{\textbf{Term}}            & \textbf{Meaning} \\
\hline
CM                       & Confusion Matrix \\
\hline
GT                       & Ground Truth \\
\hline
P                        & Prediction \\
\hline
True                     & Patch/Image contains Markush at least one structure \\
\hline
False                    & Patch/Image contains no Markush structure(s) \\
\hline
Patch-level              & Each patch is evaluated separately \\
\hline
Image-level              & The predictions for each patch in an image are combined to give a prediction for the entire image that they constitute. \\
\hline
FC Only                  & Only the fully connected layers are changed during training, other layer parameters are frozen.\\
\hline
Full Model               & All parameters in the model are trained. \\
\hline
Patch - Val F1           & Patch-level Macro F1 score on the validation part of the primary dataset \\
\hline
Patch - Test F1          & Patch-level Macro F1 score on the test part of the primary dataset \\
\hline
Image - Test F1          & Image-level Macro F1 score on the test part of the primary dataset \\
\hline
Image - Aux Test F1      & Image-level Macro F1 score on the auxiliary dataset \\
\hline
\end{tabular}
\vspace{1em}
\caption{Results glossary}
\label{table:glossary}
\end{table}

\begin{table}[tb]
\centering
\begin{subtable}{0.45\columnwidth}
\centering
\caption{R18\_IN\_FC}
\begin{tabular}{|c|c|c|}
\hline
   & P: True & P: False \\
\hline
GT: True & 808 & 74 \\
\hline
GT: False & 161 & 147 \\
\hline
\end{tabular}
\end{subtable}
\hspace{0.05\columnwidth}
\begin{subtable}{0.45\columnwidth}
\centering
\caption{R18\_IN\_FM}
\begin{tabular}{|c|c|c|}
\hline
   & P: True & P: False \\
\hline
GT: True & 906 & 40 \\
\hline
GT: False & 45 & 203 \\
\hline
\end{tabular}
\end{subtable}

\vspace{1em}

\begin{subtable}{0.45\columnwidth}
\centering
\caption{IV3\_USPTO\_FC}
\begin{tabular}{|c|c|c|}
\hline
   & P: True & P: False \\
\hline
GT: True & 625 & 28 \\
\hline
GT: False & 311 & 52 \\
\hline
\end{tabular}
\end{subtable}
\hspace{0.05\columnwidth}
\begin{subtable}{0.45\columnwidth}
\centering
\caption{IV3\_USPTO\_FM}
\begin{tabular}{|c|c|c|}
\hline
   & P: True & P: False \\
\hline
GT: True & 922 & 12 \\
\hline
GT: False & 114 & 104 \\
\hline
\end{tabular}
\end{subtable}

\vspace{1em}

\begin{subtable}{0.45\columnwidth}
\centering
\caption{IV3\_IN\_FC}
\begin{tabular}{|c|c|c|}
\hline
   & P: True & P: False \\
\hline
GT: True & 690 & 89 \\
\hline
GT: False & 99 & 118 \\
\hline
\end{tabular}
\end{subtable}
\hspace{0.05\columnwidth}
\begin{subtable}{0.45\columnwidth}
\centering
\caption{IV3\_IN\_FM}
\begin{tabular}{|c|c|c|}
\hline
   & P: True & P: False \\
\hline
GT: True & 742 & 10 \\
\hline
GT: False & 64 & 260 \\
\hline
\end{tabular}
\end{subtable}
\vspace{1em}
\caption{Confusion matrices of the 6 CNN configurations for the patch-level primary test set results. The experiments were run 5 times, these confusion matrices are from the first of the 5 runs.}
\label{table:confusion_matrices}
\end{table}

The main results of the experiments can be seen in \autoref{table:main_results}. A glossary explaining the terminology used can be seen in \autoref {table:glossary}. The most informative column in \autoref{table:main_results} is the ``Patch - Test F1" column. This is how the model performed on unseen data from the primary dataset at the optimized patch level, and will be used for most comparisons. The confusion matrices for the "Patch - Test F1" evaluation are available in \autoref{table:confusion_matrices}. Results are also shown at image level, and at image level on the auxiliary dataset. No patch level results are available on the auxiliary set because it was not annotated. Because the method for computing an image label from the patch labels has not been optimized during this research, these values can be seen as a lower bound.

\subsubsection{Baseline}

The ORB baseline did not perform very well, ranking second to last of all methods tried on the Image-level Test set. Furthermore, the results were inconsistent, the performance on the auxiliary set was significantly lower than the performance on the primary test set. Because ORB is a full-image method, no patch-level results are available. On the primary test set, the baseline got a Macro F1 of 0.701 $\pm$ 0.052, and it performed worse on the much larger auxiliary set with a score of 0.533. The confusion matrix for the auxiliary dataset evaluation can be seen in \autoref{table:confusion_matrices}.

\subsubsection{Layers}

In all of the experiment configurations, the configuration in which the full model was trained performed significantly better than the equivalent configuration in which only the fully connected layers were trained. The improvement on the patch-level test set was 0.145, 0.195 and 0.20 Macro F1 score. These values are well outside the confidence interval and thus are significant at a 95\% confidence level.

A reason for this improvement could be that the filters learned by pretraining are not well fit for discerning Markush structures, and can thus be improved during the fine-tuning. For the ImageNet models this would be a natural assumption, since ImageNet contains no images that look like Markush structures. However, it is not clear why the features of the USPTO-pretrained models still required their filters to be fine-tuned.

\subsubsection{Architecture}

With respect to architecture, the results are less clear. We can compare models, keeping pretraining (ImageNet) and Layers trained (FC only and Full Model) equal. In the FC only case, IV3\_IN\_FC and R18\_IN\_FC's confidence intervals overlap, meaning there is no significant difference. In the Full Model case, IV3\_IN\_FM has a Macro F1 of 0.028 higher than R18\_IN\_FM. Although this is a small difference relative to the other scores, both of these models performed very reliably and thus their confidence intervals are small enough not to overlap. We can say that at 95\% confidence level that InceptionV3 performs better than ResNet18 at this task.

A possible explanation for this can be found in the input sizes. Because InceptionV3 has a slightly larger input size, it sees more context around any Markush indicators, which can help with classification. Furthermore, the model has twice as many parameters, which could help with classification as well. However, this seems a less likely explanation because of the simplicity of the data. More research would be required to answer this question fully. 

\subsubsection{Pretraining}

With regard to pretraining, we can compare the InceptionV3 models on the Patch - Test F1 score. In both the FC only and Full Model configurations the ImageNet pre-trained models outperformed the USPTO-pretrained models. By 0.142 F1 (FC) and 0.158 (Full Model). 

This was unexpected, as the ImageNet images are in a completely different domain than the USPTO images. However, the ImageNet dataset is much larger, and very complex. A possible explanation is that the filters learned by ImageNet pretraining are so robust that they are also applicable outside the ImageNet domain. Why the filters learned by the USPTO set underperform is unclear, more investigation into Elsevier's proprietary model would be required. 

\subsection{ROC Curves}

\begin{figure}[tb]
    \centering
    \includegraphics[width=0.4\textwidth]{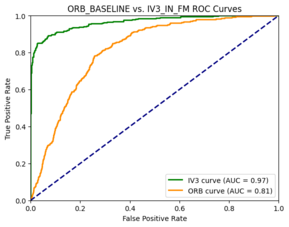}
   \caption{The ROC curves for the ORB\_BASELINE model as evaluated on the auxiliary dataset, and for the IV3\_IN\_FM on the primary patch test set.}
    \label{fig:TwoROCCurves} 
\end{figure}

The receiver operating characteristic (ROC) curves for the baseline and best CNN configuration are shown in \autoref{fig:TwoROCCurves}. These two ROC curves can not be compared directly because they were generated on different data. The ORB\_BASELINE can only give prediction probabilities for full images, so the ROC curve was created for the larger auxiliary dataset. The IV3\_IN\_FM model generates prediction probabilities for individual patches, not whole images. Therefore the ROC curve for IV3\_IN\_FM was generated based on prediction probabilities on the primary dataset's test section. Despite this incongruency, the difference in performance is so large that it is clear that the IV3\_IN\_FM's ROC is far superior to the ORB\_BASELINE's ROC. This is supported by the results from \autoref{table:main_results}, which show us that the results at patch level generally hold up at the image level. 

An interesting observation that can be made is the steepness of the IV3\_IN\_FM curve. Normally, the classification threshold for a model is set at the point which is closest to the top-left corner of the ROC curve (1.0 TPR, 0.0 FPR). However, for the IV3\_IN\_FM ROC curve, at circa 0.7 TPR, the FPR is still nearly 0.0. By leveraging this classification characteristic at patch level, the classification performance on the image level can be improved further. On average, an image in the primary dataset contained 9.52 Markush indicators (\autoref{table:Dataset_stats}). If we label an image ``Markush" if only one of its patches is labeled "Markush", this means that this gives us, on average, 9.52 opportunities to correctly identify the Markush indicators in an image. If we naively assume that the Markush indicators in the image are independent of each other and distributed evenly over the entire dataset we could theoretically get the performance as seen in \autoref{tab:naive_performance}.

\begin{table}[tb]
    \centering
    \begin{tabular}{|c|c|}
    \hline
    \# of Indicators in Image & Naïve Expected Image-TPR @ 0.7 TPR                              \\ \hline
    1                            & $0.7^1 = 0.7$                   \\ \hline
    5                            & $0.7^{5^{-1}} = 0.931$     \\ \hline
    10                           & $0.7^{{10}^{-1}} = 0.965$     \\ \hline
    \end{tabular}
    \vspace{0.2cm}
    \caption{Naive performance at image level.}
    \label{tab:naive_performance}
\end{table}

Because the Markush indicators that can be found in a single image are very much dependent on each other, this calculation should be only seen as an upper bound of the theoretical TPR. However, it does give an indication of how much performance can increase with multiple Markush indicators in a single image.

\subsection{Error Investigation}

In this section, errors that are made by the models are shown. In the ORB case, the left image is the template used and the right image is the query image. In the CNN case, the left image is the query patch, and the right image is a saliency map of the classification as done by the IV3\_IN\_FM model. This saliency map shows which pixels contributed strongest to the classification of the image, where brighter means a larger contribution.

\subsection{ORB}

\begin{figure}[tb]
    \centering
    \begin{minipage}{0.3\textwidth}
        \includegraphics[width=\textwidth]{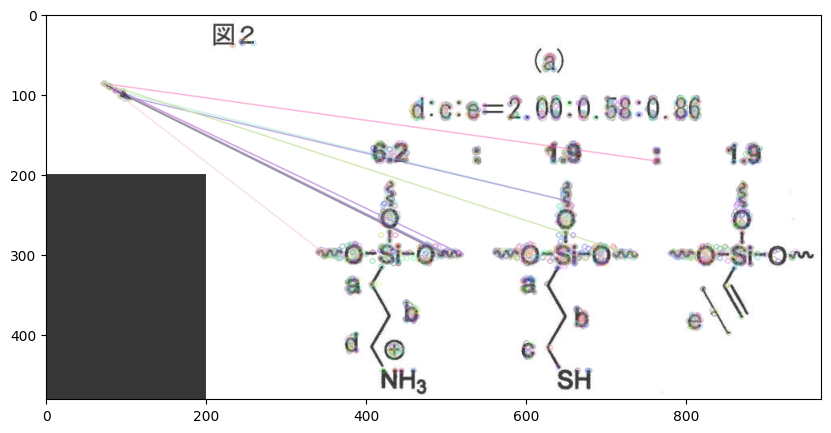}
        \caption{An ORB matching example where the template indicator is not present in the image.}
        \label{fig:orb_indicator_missing}
    \end{minipage}\hspace{2cm}
    \begin{minipage}{0.3\textwidth}
        \includegraphics[width=\textwidth]{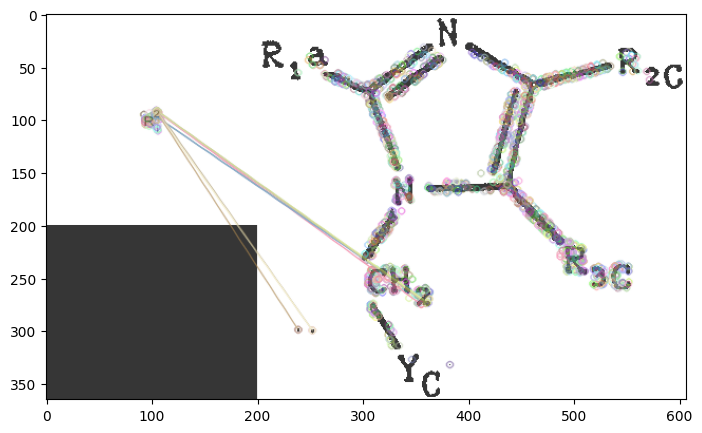}
        \caption{An ORB matching example where the template indicator differs in scale and font from the indicators in the image.}
        \label{fig:orb_scale_difference}
    \end{minipage}
\end{figure}

\autoref{fig:orb_indicator_missing} shows a common problem with the ORB classification method. In this case, the template that was selected from the training data is an arrow. This is a relatively rare Markush indicator, and is not used in the query image. Because of this, no good matches are possible. This problem is alleviated slightly by the used method of comparing with many different Markush indicators for each query image. However, for rare Markush indicators this means that for example only 1 of the 100 templates used is able to provide a proper match, which gives obvious SNR issues.

\autoref{fig:orb_scale_difference} shows another common problem because ORB does not perform well on scale variance. In this comparison, the template image contains a small ``R". Even though the query image contains multiple indicators that are also ``R"s, they are not matched because they are much larger pixel-wise making the low-level features that ORB detects different and thus not creating any matches. Furthermore, the fonts used for the ``R"s are different, which is another problem for the relatively fragile feature matching from ORB.

\subsection{CNN} \label{sec:CNN_errors}

\begin{figure}[b]
    \centering
    \begin{minipage}{0.3\textwidth}
        \includegraphics[width=\textwidth]{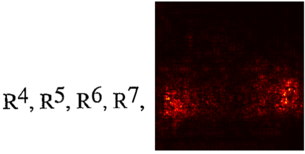}
        \caption{False Positive: A patch showing text containing many symbols that could be Markush structures, but are not in a SCF.}
        \label{fig:FP_Markush1}
    \end{minipage}\hspace{2cm}
    \begin{minipage}{0.3\textwidth}
        \includegraphics[width=\textwidth]{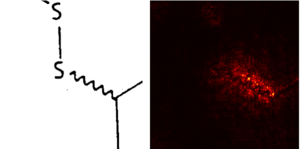}
        \caption{False Positive: a patch showing a ``wavy line", often used as a Markush indicator when not connected on both sides.}
        \label{fig:FP_Markush2}
    \end{minipage}
\end{figure}

\autoref{fig:FP_Markush1} shows a false positive patch classification. This image contains text related to Markush structures, but does not contain an actual CSF with Markush indicators and should thus be labeled False. The ``R"s are very common Markush indicators, and comma's alone can sometimes be an indication of a Markush structure, the combination of these 2 factors likely led to the false positive classification. Interestingly, the saliency map shows us that the Rs on the edges of the ``structure" were most important for the classification, while the almost exact same Rs in the middle barely contributed to the classification. In Markush structures with ``R" indicators, their location is always on the edge of a CSF. It is likely that the model saw this pattern in this text, mistaking the ``$R^5$" and ``$R^6$" for other CSF lines. This problem can likely be solved with more training data.

\autoref{fig:FP_Markush1} shows a false positive example with text containing symbols resembling Markush structures. On the other hand, \autoref{fig:FP_Markush2} illustrates another false positive where the ``wavy line" is the primary classification indicator. This line is commonly used as a Markush indicator when not connected on both sides.

\autoref{fig:FP_Markush2} shows another false positive. In this case, the saliency map shows that the ``wavy line" is mostly responsible for the classification. The ``wavy line" is a very common Markush indicator, but only when it is connected on a single side. In this case it is a special way of showing a bond between two atoms. This is a difficult problem for the CNN to solve, because it requires not only recognition of a certain structure, but also a good understanding of the context in which it is placed. This problem can likely also be solved with (much) more training data.

\begin{figure}[tb]
    \centering
    \begin{minipage}{0.3\textwidth}
        \includegraphics[width=\textwidth]{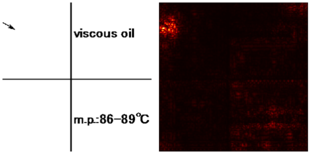}
        \caption{False Negative: the Markush Indicator is only partially visible in the patch.}
        \label{fig:FN_Markush1}
    \end{minipage}\hspace{2cm}
    \begin{minipage}{0.3\textwidth}
        \includegraphics[width=\textwidth]{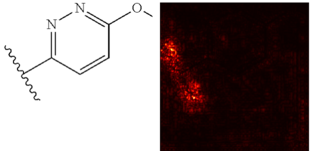}
        \caption{False Negative: a clearly visible Markush indicator is not detected.}
        \label{fig:FN_Markush2}
    \end{minipage}
\end{figure}

In \autoref{fig:FN_Markush1} we encounter a different kind of error. This error is caused by Markush indicators that are only partially included in a patch. In this case, the patch should have been labeled "True" due to the arrow (Markush indicator) in the top-left, which is only partially visible in the patch. This is caused by a fundamental problem of the method used. If all but a few of a Markush indicator's pixels fall inside of a patch, we want to label the patch ``True". If it is labeled ``False", the data would be polluted: a patch clearly containing a Markush structure would be labeled false. However, the opposite is also true. If a patch contains only a couple of pixels of an annotation, it should not be labeled ``True", this would cause the model to associate non-Markush indicators with a ``True" label. As explained in \autoref{sec:methodology}, the method used labels a patch ``True" if at least 50\% of an indicator's pixels lie inside the patch, a compromise between false positives and false negatives.

Lastly, \autoref{fig:FN_Markush2} shows a false negative with a clear Markush indicator (``wavy line") that is completely visible in the patch. The saliency map shows that the pixels around the indicator were most important for the classification. However, the patch was still labeled ``False". It is unclear why this happened. More investigation would be required to determine what caused this error.
\section{Discussion}
\label{sec:discussion}

Because the existing literature does not describe a method that achieves the same goal as this research, no direct comparison of performance can be made to another study. Because Markush structures in scientific publications have posed a significant problem for OCSR pipelines \cite{xu2022molminer}, the method proposed in this research is a valuable addition to the existing literature. With the best model having Macro F1 scores of above 0.90 at both the patch- and image-level, as well as on the larger auxiliary dataset. It can be concluded that this model performs consistently well at identifying images containing Markush structures. Furthermore, the performance can be increased even more by optimizing the image-level predictions.

\subsection{Limitations}
The results from this research do not come without limitations. The small dataset meant that performance using this method could be better than currently found. However, because it is very costly to label such data it was not possible to use a larger set. Because of the cost, the domain expert was not able to annotate each image, but merely label them. This meant that annotation was done by someone with no significant formal knowledge of chemistry. While the researcher and domain expert created an extensive  annotation guide document to aid in annotation, it can not be guaranteed that no mistakes were made during annotation, leading to incorrect training and classifications.

Furthermore, the experimental setup had some issues. Because the two architectures compared have different input sizes, the InceptionV3 model got slightly larger patches than ResNet18 (299px vs 224px squares). This meant that it can not be determined whether the input size or the rest of the architecture is responsible for the difference in performance. Moreover, each model was tested on an auxiliary dataset to test for generalizability. The results seem to hold up at the image level on the auxiliary dataset, but due to computational constraints it was not possible to run each experiment 5 times on the larger auxiliary dataset, meaning that no confidence interval is available for these results. Lastly, a fundamental problem with labeling the patches as shown in \autoref{fig:FN_Markush1} means that the patches will always contain some level of label noise. This can potentially be solved by converting the classification task into a regression task, but this was outside the scope of this research.
\vspace{-2mm}


\section{Conclusions}
\label{sec:conclusion}

This research focused on identifying Markush structures in images under challenging conditions. By filtering images with Markush structures in them, the accuracy of OCSR pipelines can be significantly improved. This filtering has not been done before in existing literature, and thus a new method was proposed. By adding manual annotations and splitting the image up into patches, this method overcomes the low signal-to-noise ratio and large variance in image sizes.

\subsection{Research Questions}

This research has shown that end-to-end deep convolutional neural networks can classify Markush structures with a high precision and recall. The best CNN, IV3\_IN\_FM, outperformed the fixed-feature CV method (ORB) with 0.928 macro F1 to 0.701 on the image-level primary test set, where the 0.928 is a lower bound and can still be improved. On the auxiliary test set, IV3\_IN\_FM outperformed the ORB method with 0.914 macro F1 to 0.533 macro F1. Lastly, the AUC of the ORB method's ROC was 0.81, while that of the IV3\_IN\_FM model was 0.97. Although there are limitations to this research, they are small enough that they do not severely impact the reliability of the results. Overall, it can be concluded that using end-to-end deep CNN's can greatly improve classification compared to a state-of-the-art fixed-feature method.

In both the ORB method and CNN method, the features used to determine the classification are interpretable to a certain extent. The ORB method visually shows the matches between the template image and the query image, which reveals which parts of the image contributed to the classification. Furthermore, the XGBoost model shows which of the templates had the highest importance. The issue with the ORB method is that it does not have reliable explanatory power. The results on the auxiliary set are significantly different from the results on the primary set, and visual inspection shows that many of the matches it generates appear to be spurious. 

The CNN method allows us to produce saliency maps which reveal which part of a patch contributed most to the classification. These saliency maps provide interpretability, however their trustworthiness is uncertain \cite{tomsett2020sanity}. Another source of interpretability comes inherently from the method by which image-level predictions are made. By looking at the labels given to each individual patch, it can be determined which patch(es) and thus which parts of an image caused an image to be labeled as Markush. This also has reliable explanatory power. If a patch is labeled ``Markush" in one image, if that same patch was present in another, the label would not change. Still, a CNN is mostly a black-box method. The amount of parameters and high dimensionality makes it nearly impossible to interpret. 

\subsection{Future Work}

One of the issues with the method as proposed is how patches are labeled when they contain only a part of a Markush indicator, as shown in \autoref{fig:FN_Markush1}. This problem could be solved by reformulating this as a regression task instead of a classification task. The model then predicts a probability that the patch contains a Markush structure: if an indicator is only partly in a patch, the value would be equal to the proportion of pixels that are present in the patch. Another technique that could be used in future work is detecting the scale of the image by analyzing the size of letters and chemical bonds. The images could then be rescaled such that they are all of approximately equal size. This would eliminate the "pixel scale variance" problem. In addition to this, future work on this problem could expand the dataset with images that originate not only from the USPTO, to increase the diversity of the data. With this method, OCSR pipelines can be improved, allowing for higher quality conversion of CSF's to machine-readable formats. In turn, this improves the quality and coverage of databases such as Reaxys, which can help other researchers in their work. Moreover, it makes it easier for automated information retrieval methods to assist those researchers, further accelerating scientific progress.


\bibliographystyle{plain}
\bibliography{bibliographies/references}


\onecolumn
\appendix

\section{Appendix}
\label{sec:apx:first_appendix}

\vspace{\fill}

\begin{table}[h]
\centering
    \begin{tabular}{l|l}
        \textbf{Hyperparameter} & \textbf{Options} \\
        \hline
        Number of ORB features generated & [500, \textbf{2000}] \\
        Number of template patches & [50, \textbf{100}, 250] \\
        Number of XGB estimators & [500, \textbf{1500}] \\
        XGB maximum depth & [\textbf{6}, 15] \\
    \end{tabular}
    \vspace{0.2cm}
    \caption{ORB Method hyperparameter options, optimal in bold.}
    \label{tab:orb_hyperparameters}
\end{table}

\vspace{\fill}

\begin{table}[h]
\centering
    \begin{tabular}{l|l}
        \textbf{Parameter} & \textbf{Options} \\
        \hline
        Learning rate & $[10^{-5}, 10^{-3}]$ \\
        Optimizer Options & Adam \cite{kingma2014adam} or SGD (Stochastic Gradient Descent) \\
        Image Augmentation probability & $[0, 1]$ \\
        Epoch count & $[1, 25]$ \\
    \end{tabular}
    \vspace{0.2cm}
    \caption{Hyperparameter ranges explored.}
    \label{tab:hyperparam_explored}
\end{table}

\vspace{\fill}

\begin{table}[h]
\centering
    \begin{tabular}{l|l}
        \textbf{Parameter} & \textbf{Options} \\
        \hline
        Maximum amount of epochs & 50 epochs \\
        Optuna trials & 25 trials \\
        Batch Size & 8 images \\
        Dataloader workers & 2 workers \\
    \end{tabular}
    \vspace{0.2cm}
    \caption{Optuna and Pytorch parameters.}
    \label{tab:optuna_pytorch_params}
\end{table}

\vspace{\fill}
\end{document}